\documentclass[sigconf]{acmart}
\usepackage{subfig}
\AtBeginDocument{%
  }

\setcopyright{acmlicensed}
\copyrightyear{2025}
\acmYear{2025}
\acmDOI{XXXXXXX.XXXXXXX}
\acmConference[ACM MM]{ACMMM 2025 Datasets Track}{October 27--31,
  2025}{Dublin, Ireland}
\acmISBN{978-1-4503-XXXX-X/2018/06}

\begin{document}

\title{A Spatial Relationship Aware Dataset for Robotics}

\author{Peng Wang}
\orcid{0000-0001-9895-394X}
\authornote{Corresponding author}
\affiliation{%
  \institution{Manchester Metropolitan University}
  \city{Manchester}
  \country{UK}
}
\email{P.Wang@mmu.ac.uk}

\author{Minh Huy Pham}
\orcid{1234-5678-9012}
\affiliation{%
  \institution{Independent Researcher}
  \city{Manchester}
  \country{UK}
}
\email{minhhuypham.working@gmail.com}

\author{Zhihao Guo}
\orcid{1234-5678-9012}
\affiliation{%
  \institution{Manchester Metropolitan University}
  \city{Manchester}
  \country{UK}
}
\email{ZHIHAO.GUO@stu.mmu.ac.uk}

\author{Wei Zhou}
\orcid{1234-5678-9012}
\affiliation{%
  \institution{Cardiff University}
  \city{Cardiff}
  \country{UK}
}
\email{ZhouW26@cardiff.ac.uk}

\renewcommand{\shortauthors}{Peng, Minh, Zhihao, Wei}

\begin{abstract}
Robotic task planning in real-world environments requires not only object recognition but also a nuanced understanding of spatial relationships between objects. We present a spatial-relationship-aware dataset of nearly 1,000 robot-acquired indoor images, annotated with object attributes, positions, and detailed spatial relationships. Captured using a Boston Dynamics Spot robot and labelled with a custom annotation tool, the dataset reflects complex scenarios with similar or identical objects and intricate spatial arrangements. We benchmark six state-of-the-art scene-graph generation models on this dataset, analyzing their inference speed and relational accuracy. Our results highlight significant differences in model performance and demonstrate that integrating explicit spatial relationships into foundation models, such as ChatGPT 4o, substantially improves their ability to generate executable, spatially-aware plans for robotics. The dataset and annotation tool are publicly available at \url{https://github.com/PengPaulWang/SpatialAwareRobotDataset}, supporting further research in spatial reasoning for robotics.
\end{abstract}

\ccsdesc[500]{Computing methodologies~Robotic planning}
\ccsdesc[500]{Computing methodologies~Vision for robotics}


\keywords{Scene-graph generation, Spatial relationships, Robotics, Dataset, Annotation tool, Object detection, Relation prediction
}
\begin{teaserfigure}
  \includegraphics[width=\textwidth]{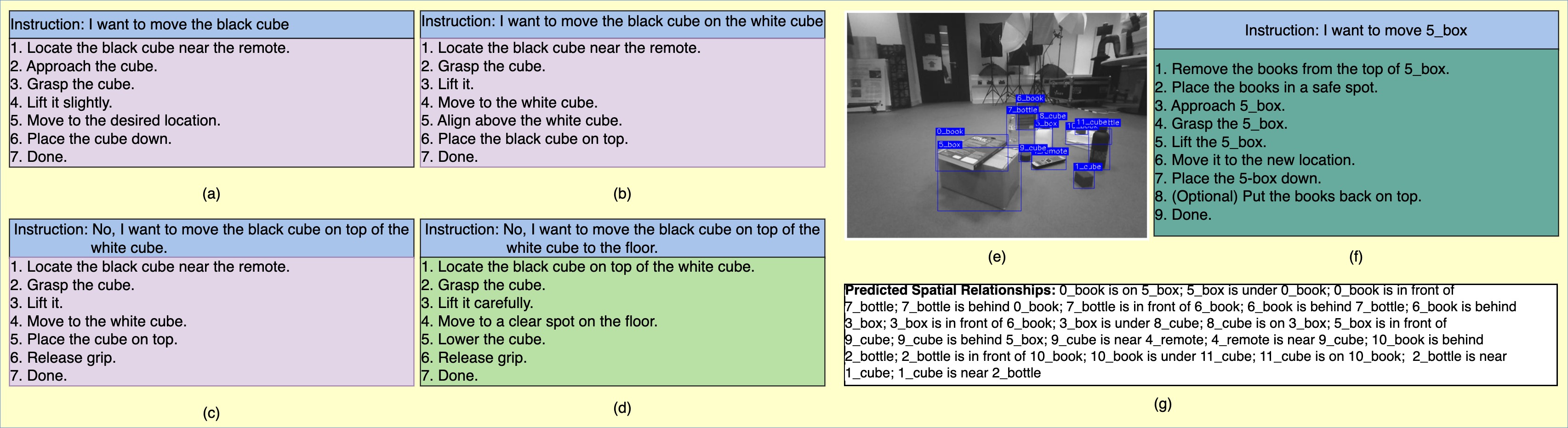}
  \caption{Considering the spatial relationships between objects is
    crucial for robotic tasks. In this work, we present a custom
    dataset of robot-acquired images annotated with spatial
    relationships, positions and object attributes, and evaluate the performance of six SOTA
    scene-graph generation models on this dataset.  While our findings reveal
    significant differences in inference speed and relational accuracy of different models on our dataset, we find that integrating spatial relationships into foundation models such as ChatGPT 4o can significantly improve their performance on robotic task planning. \textbf{Note} numbers in e, f, and g can be replaced by attributes. The dataset is available at \url{https://github.com/PengPaulWang/SpatialAwareRobotDataset}}
  \Description{
    A robot-acquired image annotated with spatial relationships. The
    image shows various objects arranged in a laboratory environment,
    with bounding boxes around each object and lines connecting them to
    indicate their spatial relationships. The relationships include
    "on", "in front of", "behind", "to the left of", "to the right of",
    "under", and "near". The image is used to evaluate the performance
    of scene-graph generation models.
  }
  \label{fig:teaser}
\end{teaserfigure}

\received{20 February 2007}
\received[revised]{12 March 2009}
\received[accepted]{5 June 2009}

\maketitle

\section{Introduction}

Robotic task planning in real-world environments requires not only object recognition but also a deep understanding of spatial relationships between objects. While recent advances in Large Language Models (LLMs)~\cite{touvron2023llama} and Vision-Language Models (VLMs)~\cite{radford2021learning} have enabled embodied agents like robots to interpret natural language instructions and generate action plans~\cite{wang2024llm,xiao2023llm}, these models often overlook the critical role of spatial reasoning. This limitation is especially pronounced in robotics, where the physical arrangement of objects can determine the success or failure of a task.

For example, as illustrated in Figure~\ref{fig:teaser}, the state-of-the-art VLM model, ChatGPT 4o, was presented with an image (Figure~\ref{fig:teaser}e, without annotations) and prompted with the instruction "I (as in robot) want to move the book" (Figure~\ref{fig:teaser}a). The intended book is the white book under a black sponge cube in the background. ChatGPT 4o is expected to understand the image and instruction, then plan a series of rational actions for the robot to complete the task. The challenges are twofold: (1) There are several books in the image, so ChatGPT 4o must identify the correct one; (2) The intended book is under a sponge cube, so the robot needs to remove the cube before moving the book. This type of spatial relationship awareness is essential in real life, as some objects (e.g., dishes) are fragile and must be removed before moving items beneath them. As shown in Figure~\ref{fig:teaser}a--d, despite ChatGPT 4o's strengths in commonsense reasoning, it requires explicit spatial details to generate executable, spatially-aware plans for robots. For instance, neither Figure~\ref{fig:teaser}a nor b meets the requirements. Figure~\ref{fig:teaser}c provides a proper plan, but a slight change in the instruction (adding "onto the floor") reverts it to an unexecutable plan. This reveals how sensitive LLMs and VLMs are to instruction details, and how easily they can generate unexecutable plans, even for SOTA models like ChatGPT 4o.

Recent works have addressed the challenge of spatial understanding in robotics by leveraging VLMs and large-scale spatial datasets. Cai et al.~\cite{cai2024spatialbot} introduced SpatialBot, which enhances spatial reasoning by incorporating both RGB and depth images. However, reliance on depth images limits applicability in scenarios where such data is unavailable. Similarly, Song et al.~\cite{song2025robospatial} presented ROBOSPATIAL, a large-scale dataset of real indoor and tabletop scenes annotated with rich spatial information, enabling improved spatial reasoning and manipulation tasks in robotics. This dataset, however, depends on 3D scans for spatial annotations.

In contrast, our approach focuses on a dataset acquired directly by robots using only RGB images, capturing scenes with identical or similar objects and intricate spatial relationships (e.g., one object on top of another). Spatial reasoning is achieved through scene-graph generation (SGG), rather than relying on depth or 3D scans. Notably, Neau et al.~\cite{neau2024react} have also explored the impact of spatial relationships in robotic tasks, but their work primarily uses the Gnome dataset, which is not tailored for robotics. The proposed dataset is annotated with object attributes, positions, and spatial relationships. It reflects the complexities of real-world robotic scenarios, including scenes with similar or identical objects and intricate spatial arrangements. Images were collected using a Boston Dynamics Spot robot and annotated with a purpose-built tool that streamlines the labelling of spatial relationships, positions, and object attributes. The annotated dataset was used to train and evaluate six SOTA SGG models, assessing their inference speed and relational accuracy, and ultimately integrating spatial relationship information into foundation models such as ChatGPT 4o to enhance their effectiveness in robotic task planning. For instance, as shown in Figure~\ref{fig:teaser}f, providing SGG-derived spatial relationships to ChatGPT 4o enables it to generate executable, spatially-aware plans.

The main contributions of this work are: (1) the creation of a spatial-relationship-aware dataset to enable robotic task planning in scenes with complex spatial relationships and identical objects; (2) a comprehensive evaluation of leading SGG models on this dataset; and (3) evidence that integrating spatial relationships into LLM/VLM-based planning improves real-world task execution.

\section{Related Work}
SGG has emerged as a pivotal technique in robotic perception, enabling structured understanding of environments by modelling both objects and their spatial relationships. This structured representation is essential for robotic tasks such as object manipulation, navigation, and task planning, where spatial reasoning is critical for success~\cite{neau2024react}.

Traditional SGG research has focused on large-scale, diverse datasets such as Visual Genome~\cite{krishna2017visual}, which emphasise general scene understanding. However, these datasets often lack the complexity and specificity required for real-world robotic applications, where scenes may contain multiple similar or identical objects and intricate spatial arrangements. Recent advances in LLMs/VLMs~\cite{touvron2023llama, radford2021learning,wu2023embodied,vemprala2023chatgpt} have improved the ability of embodied agents to interpret instructions, but spatial reasoning remains a significant challenge, as highlighted in our introduction.

SGG typically comprises several sub-tasks, including Predicate Classification (PredCls), Scene Graph Classification (SGCls), and Scene Graph Detection (SGDet)~\cite{neau2024react,he2024g}. Our work focuses on SGDet, which involves detecting objects and predicting the predicates that connect each ordered pair of detected objects. This is particularly relevant for robotics, where accurate detection and relational reasoning directly impact task execution.

To address the limitations of existing datasets and evaluate SGG models in realistic robotic contexts, we introduce a custom dataset of robot-acquired indoor images, annotated with position, object attributes and spatial relationships. Our evaluation pipeline consists of a two-stage approach: first, we employ a YOLOv10m backbone for efficient and accurate object detection; second, we attach one of six SOTA relation-prediction heads, including Prototype-based Embedding Network (PE-NET)~\cite{zheng2023pe-net}, VCTree Predictor~\cite{tang2018vctree}, REACT Predictor~\cite{neau2024react}, Motif Predictor (Stacked Motif Networks)~\cite{zellers2018motif}, Causal Analysis Predictor (Unbiased Causal Total Direct Effect)~\cite{tang2020causal}, and Transformer Predictor~\cite{zhang2017trans}.

We benchmark each model's inference speed and relational accuracy using standard Recall@K (R@20, R@50, R@100) and mean Recall@K (mR@20, mR@50, mR@100) metrics. R@K measures the fraction of ground-truth relations recovered among the model’s top-K predictions, while mR@K averages per-predicate recall to account for class imbalance. Object detection performance is reported via mAP@50.

By evaluating these models on our spatial-relationship-aware dataset, we provide insights into their strengths and limitations in robotic task planning scenarios and demonstrate the importance of explicit spatial reasoning for improving the effectiveness of foundation models in robotics.

\section{Dataset and Annotation Tool}

\subsection{Data Collection}

Our dataset currently consists of nearly 1,000 images featuring bottles, remotes, cubes, and other objects arranged in diverse configurations within a laboratory environment. We plan to continue expanding the dataset by adding images captured under varying conditions. All images were collected using a Boston Dynamics Spot robot, providing realistic perspectives and object arrangements relevant to robotics applications.

\begin{figure*}[hbpt]
    \centering
    {\includegraphics[width=0.24\linewidth]{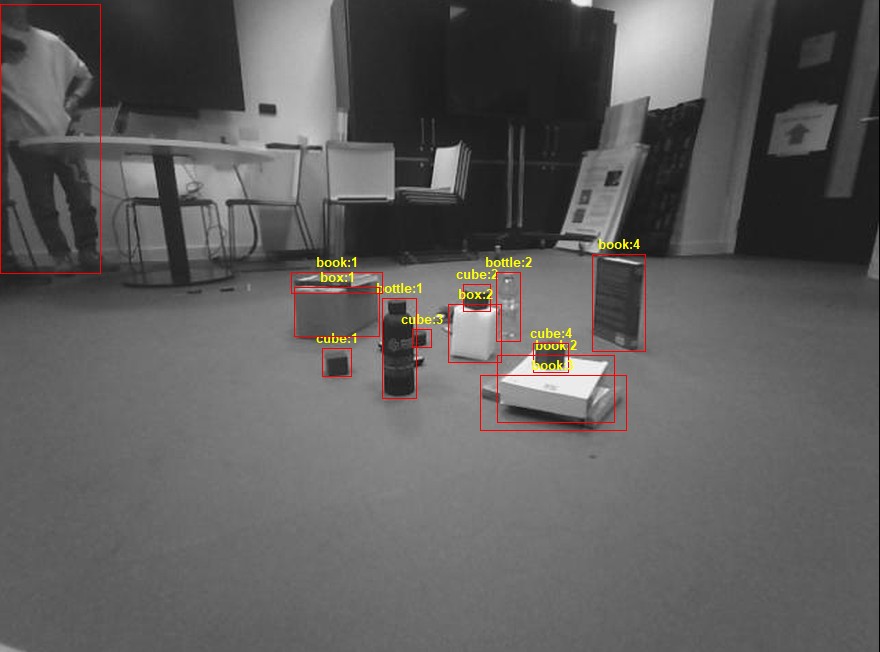}}
    {\includegraphics[width=0.24\linewidth]{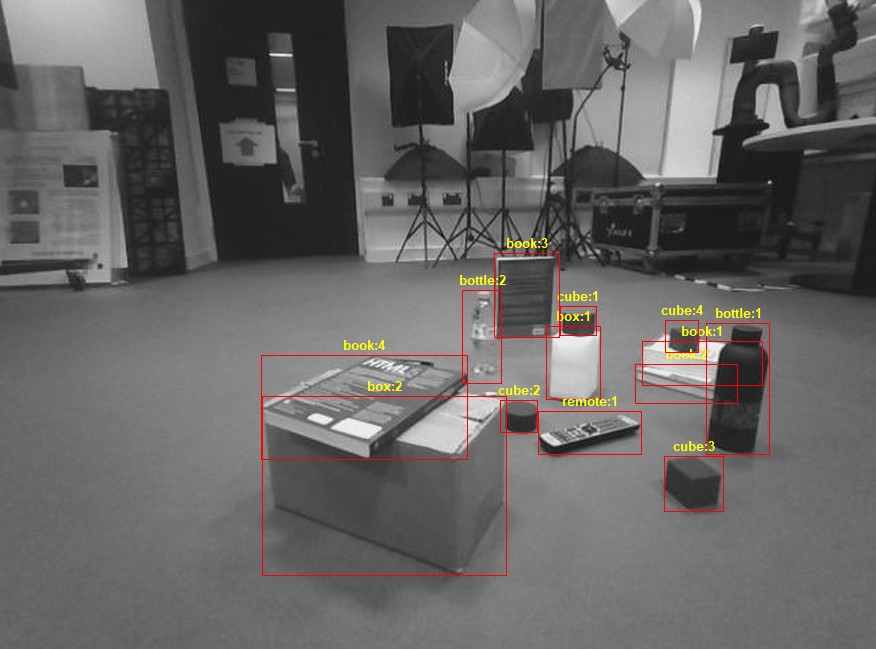}}
    {\includegraphics[width=0.24\linewidth]{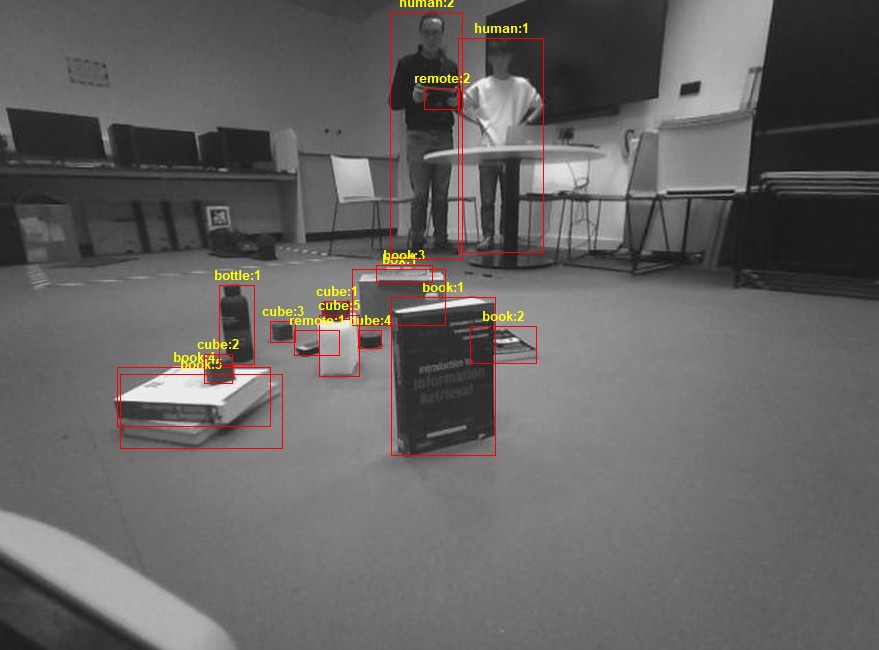}}
    {\includegraphics[width=0.24\linewidth]{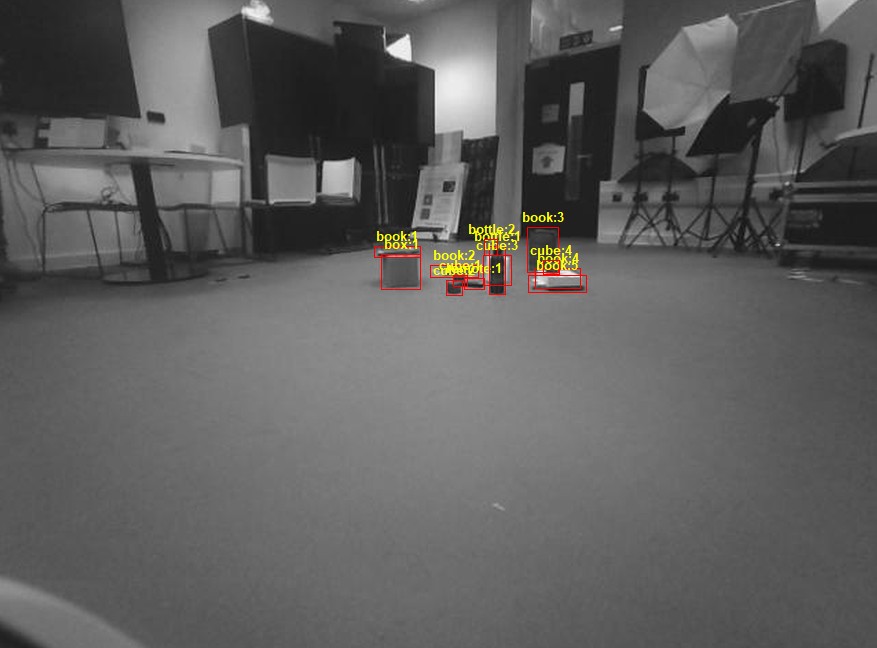}}
    \caption{Illustrative examples of spatial aware dataset for robotics (Please zoom in for the annotations). The dataset features multiple similar/identical objects in the scene with intricate spatial relationships. The data were annotated with bounding boxes, attributes, and spatial relationships. The annotations include relationships such as "on", "in front of", "behind", "to the left of", "to the right of", "under", and "near", etc. These images are used to evaluate the performance of SGG models in robotic contexts. The data were annotated using a custom-built annotation tool, which allows annotators to draw bounding boxes around objects and assign attributes to the objects and spatial relationships between them. The annotations are saved in Visual Genome format, which can be directly used for training and evaluation of SGG models. The interface of the annotation tool is shown in Figure~\ref{fig:annotation_tool}.}
    \label{fig:datasetexample}
\end{figure*}

\subsection{Annotation Process}

To provide a benchmark for evaluating SGG models in robotics contexts, we annotated each image with both object attributes and spatial relationships, rather than relying on off-the-shelf scene-graph tools, which often assume large, diverse corpora and pixel-level segmentation. We developed a lightweight bounding-box annotation tool tailored for our needs. Annotators loaded each 640x480 robot-captured image, drew axis-aligned bounding boxes around every object, and assigned object classes and predicates by clicking from "subject" to "object" to specify the relationship.

The annotation tool exported labels directly to YOLOv10m-style .txt files for object detection training, and all subject-predicate-object triplets were saved in Visual Genome format for seamless integration with relation prediction models.

\subsection{Object and Relationship Vocabulary}

Within our laboratory environment, we established a fixed vocabulary of object categories: various books, plastic and metal bottles, foam cubes, remote controls, and humans. The initial relationship list consisted of eighteen spatial predicates (e.g., "above," "behind," "part of," "mounted on," "far away from," "near," "on," "under," "to the left of," "to the right of"). However, pilot annotation revealed severe class imbalance and overlap between similar terms. To address this, we consolidated the predicate list to seven well-defined relations: behind, in front of, on, to the left of, to the right of, under, and near, ensuring each appeared in sufficient quantity and could be unambiguously interpreted.


\subsection{Annotation Workflow and Quality Control}

Nine trained annotators worked independently in batches of 100 images, drawing bounding boxes and specifying subject-predicate-object triplets. Annotators received training on the tool and were provided with precise definitions for each predicate to ensure consistency. However, during annotation, we still observed inconsistencies, particularly with the "near" predicate, and identified some images containing fewer than two objects, which cannot form valid relations. To address these issues, we conducted a centralised cleaning pass: under-annotated images were discarded, and conflicting labels were resolved by majority vote among annotators.

\begin{figure*}[ht]
  \centering
  \subfloat[Before predicates cleaning\label{fig:beforecleaning}]{
    \includegraphics[width=0.48\linewidth]{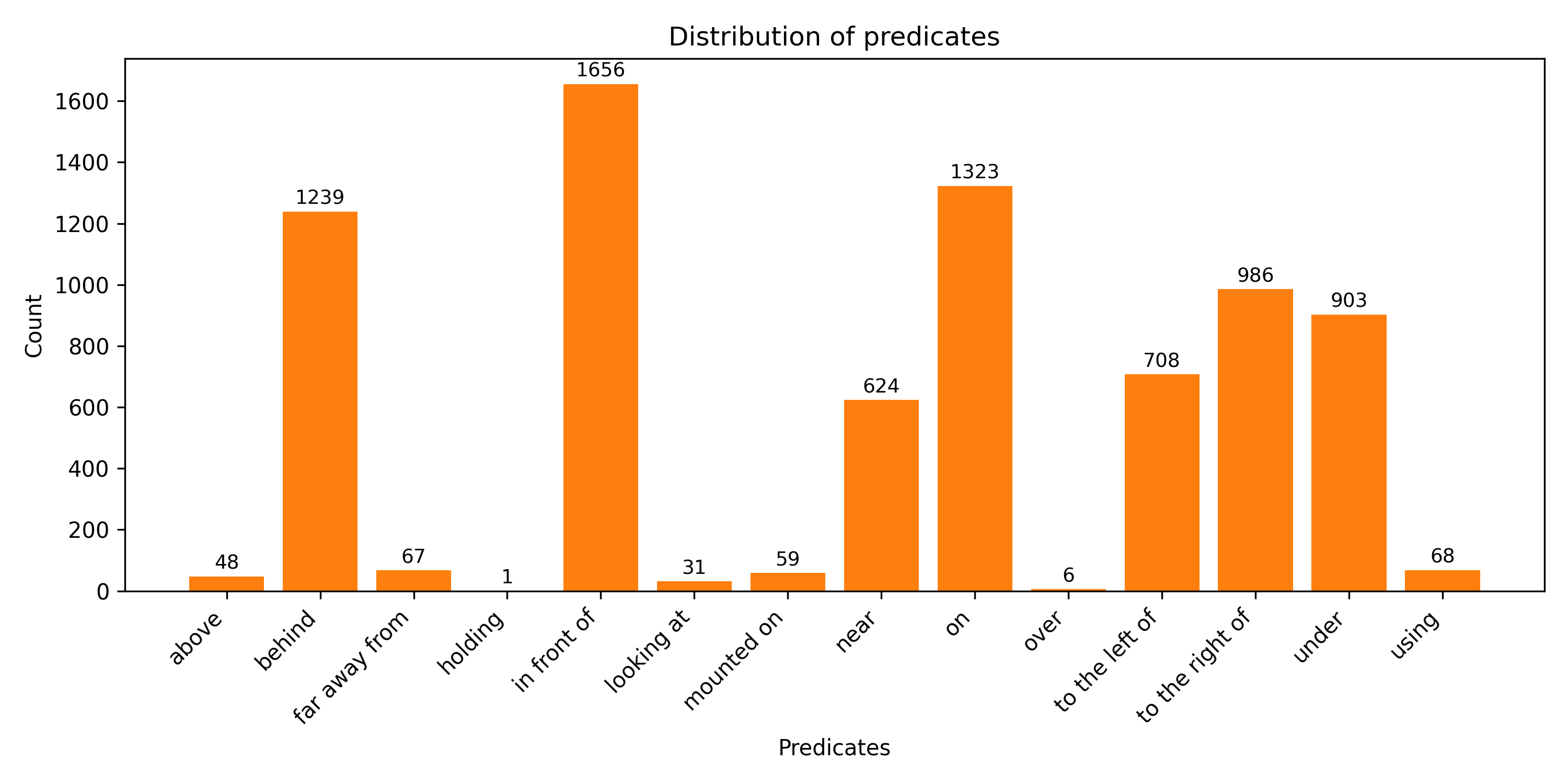}
  }
  \hfill
  \subfloat[After predicates cleaning\label{fig:aftercleaning}]{
    \includegraphics[width=0.48\linewidth]{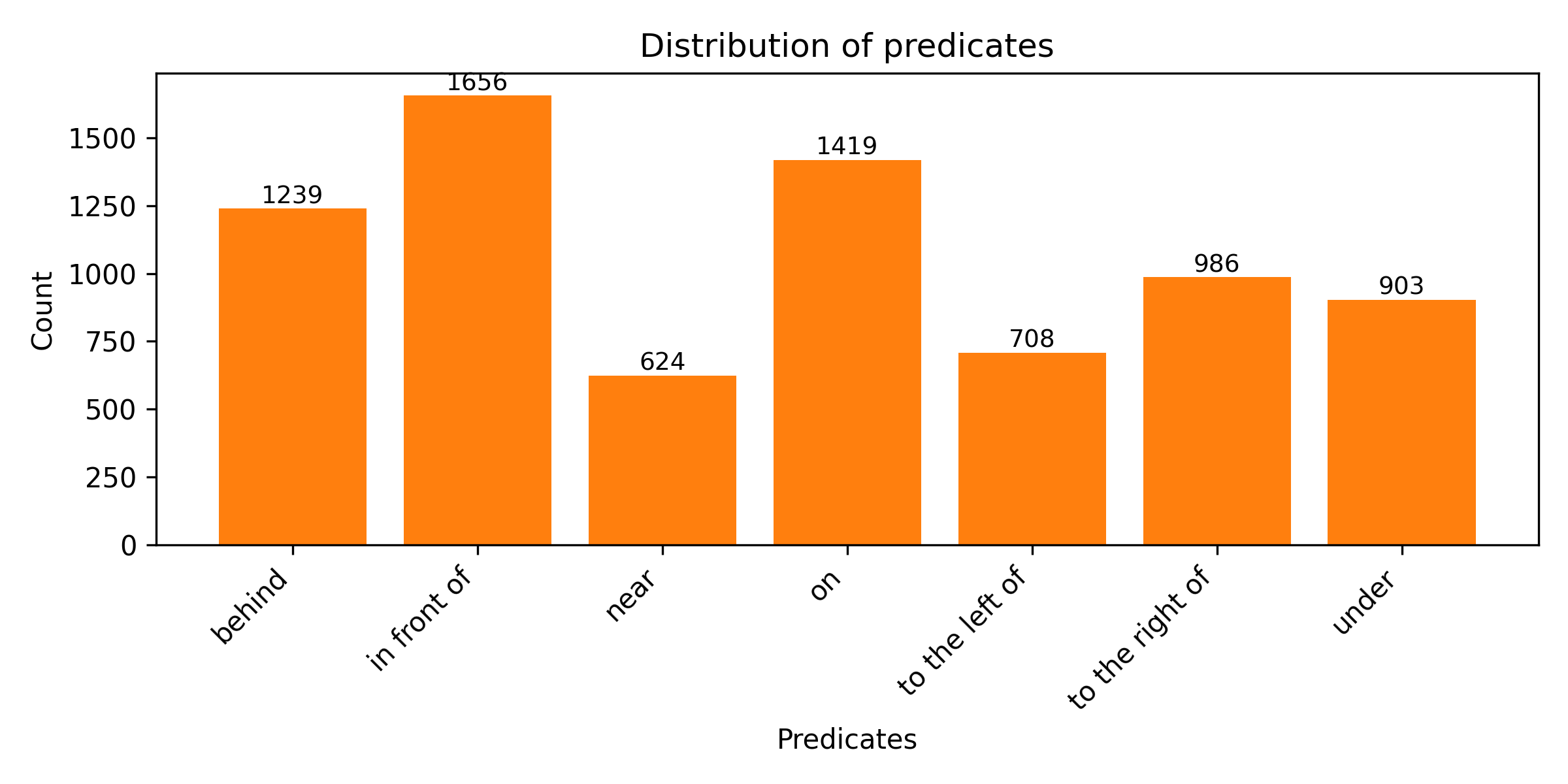}
  }
  \\
  \subfloat[Final label distribution\label{fig:labelsdistribution}]{
    \includegraphics[width=0.48\linewidth]{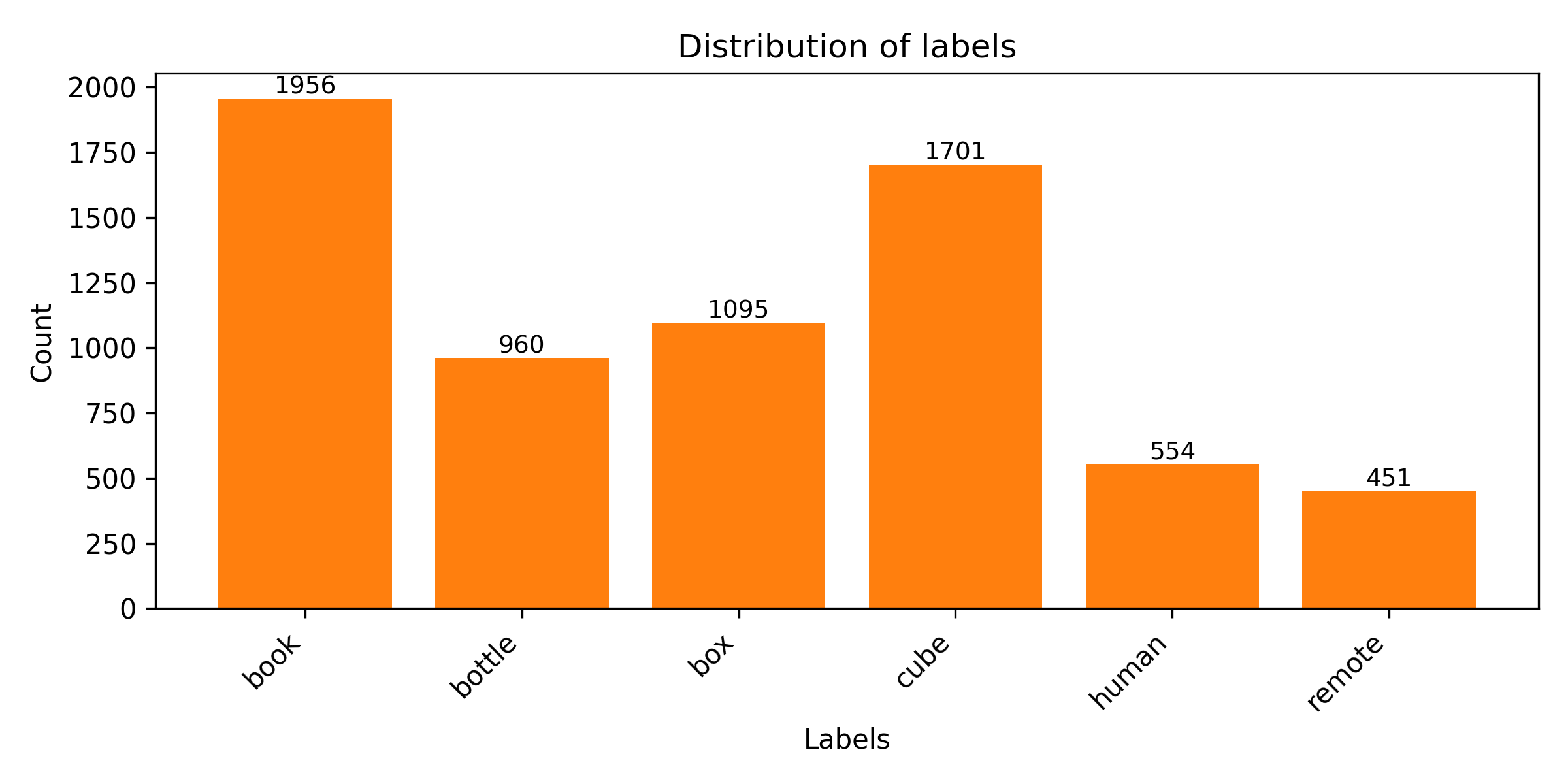}
  }
  \hfill
  \subfloat[Final attributes distribution\label{fig:predicatesdistribution}]{
    \includegraphics[width=0.48\linewidth]{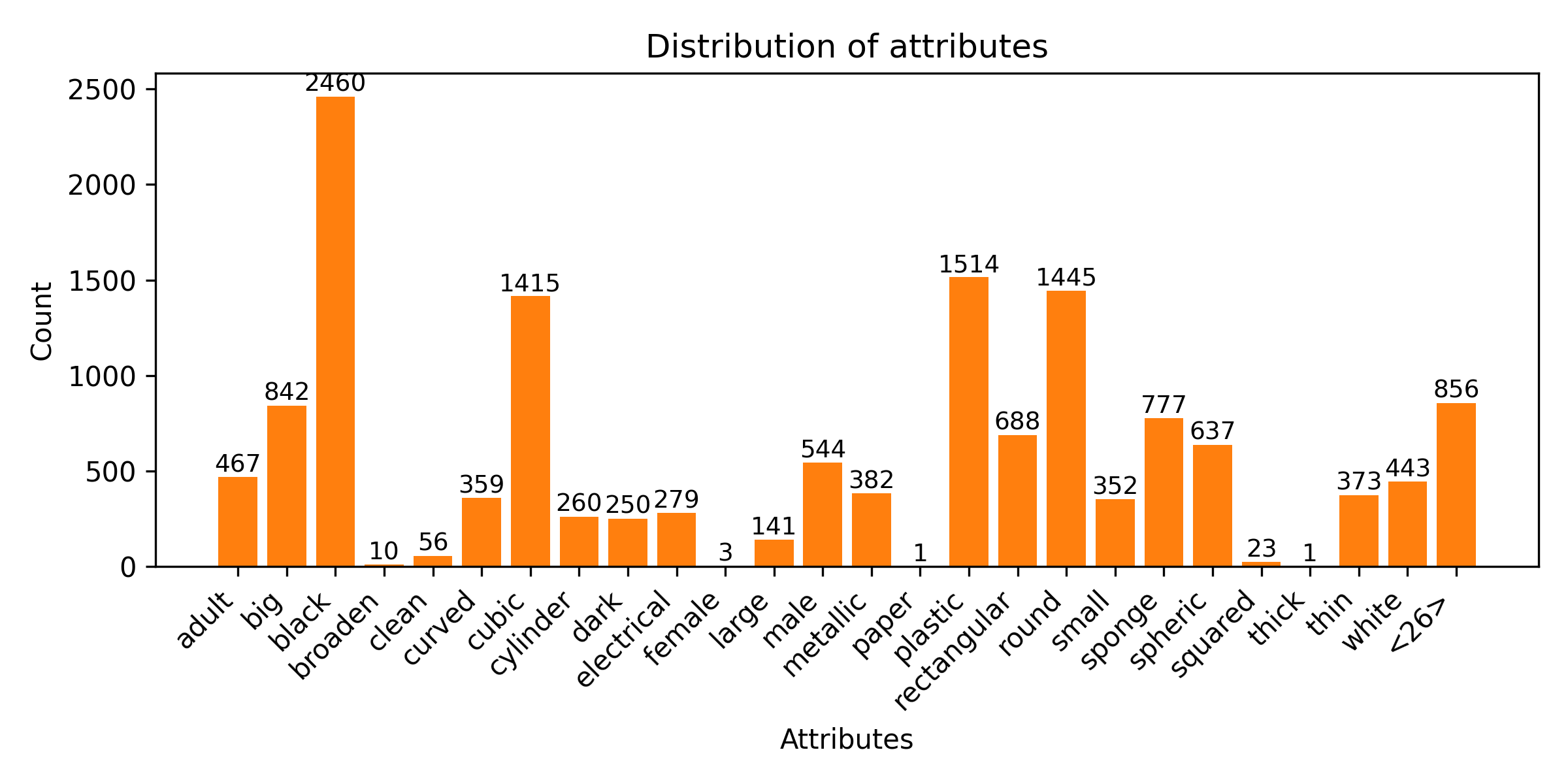}
  }
  \caption{Distributions of relationships and labels in our custom dataset. (a) and (b) show the predicate distributions before and after cleaning, respectively. The cleaning process removed rare predicates and merged similar ones into a single category. (c) shows the distribution of object labels, while (d) shows the distribution of attributes assigned to objects in the final dataset.}
  \label{fig:distributions}
\end{figure*}

Figures~\ref{fig:beforecleaning} and~\ref{fig:aftercleaning} show the predicate distributions before and after cleaning. In addition to removing rare predicates such as `holding', we merged similar predicates (e.g., `above', `over', and `on') into a single category (`on'). While subtle differences exist among these terms, annotators often interpreted them inconsistently, making them difficult to distinguish in practice. There could be hidden insights around the different interpretations of similar predicates,  despite of a precise description, which will be explored in our future works. 

The resulting validation set comprises approximately 900 robot-acquired images, each depicting 5–10 objects and annotated with an average of ten spatial relationships. Figure~\ref{fig:datasetexample} presents four representative scenes with final bounding-box and relationship annotations. Figure~\ref{fig:labelsdistribution} and Figure~\ref{fig:predicatesdistribution} show the final label and predicates distribution. Although some imbalance remains, the dataset were proven to work in the SGG processes. Future work will focus on expanding the dataset with more diverse scenes and refining the annotation process to further improve quality and consistency.

\subsection{Annotation Tool}
To facilitate efficient and consistent annotation, we developed a custom tool, SGDET-Annotate, specifically tailored for spatial relationship labelling in robotics datasets. The tool provides an intuitive interface for annotators to load image sets, define object classes, spatial predicates, and attributes, and perform all annotation tasks within a unified workflow. Users can create and adjust axis-aligned bounding boxes, assign object categories, specify subject-predicate-object relationships through guided selection, and annotate object attributes. The interface enforces annotation integrity by preventing duplicate triplets and limiting attribute assignments per object. Upon completion, annotations are exported simultaneously in Visual Genome JSON format (with bounding boxes scaled to multiple resolutions and all relationships/attributes included) and YOLO-style text files for object detection training. This streamlined process ensures high-quality, standardised data and accelerates the benchmarking of scene-graph generation models.
\begin{figure}
    \centering
    \includegraphics[width=1.0\linewidth]{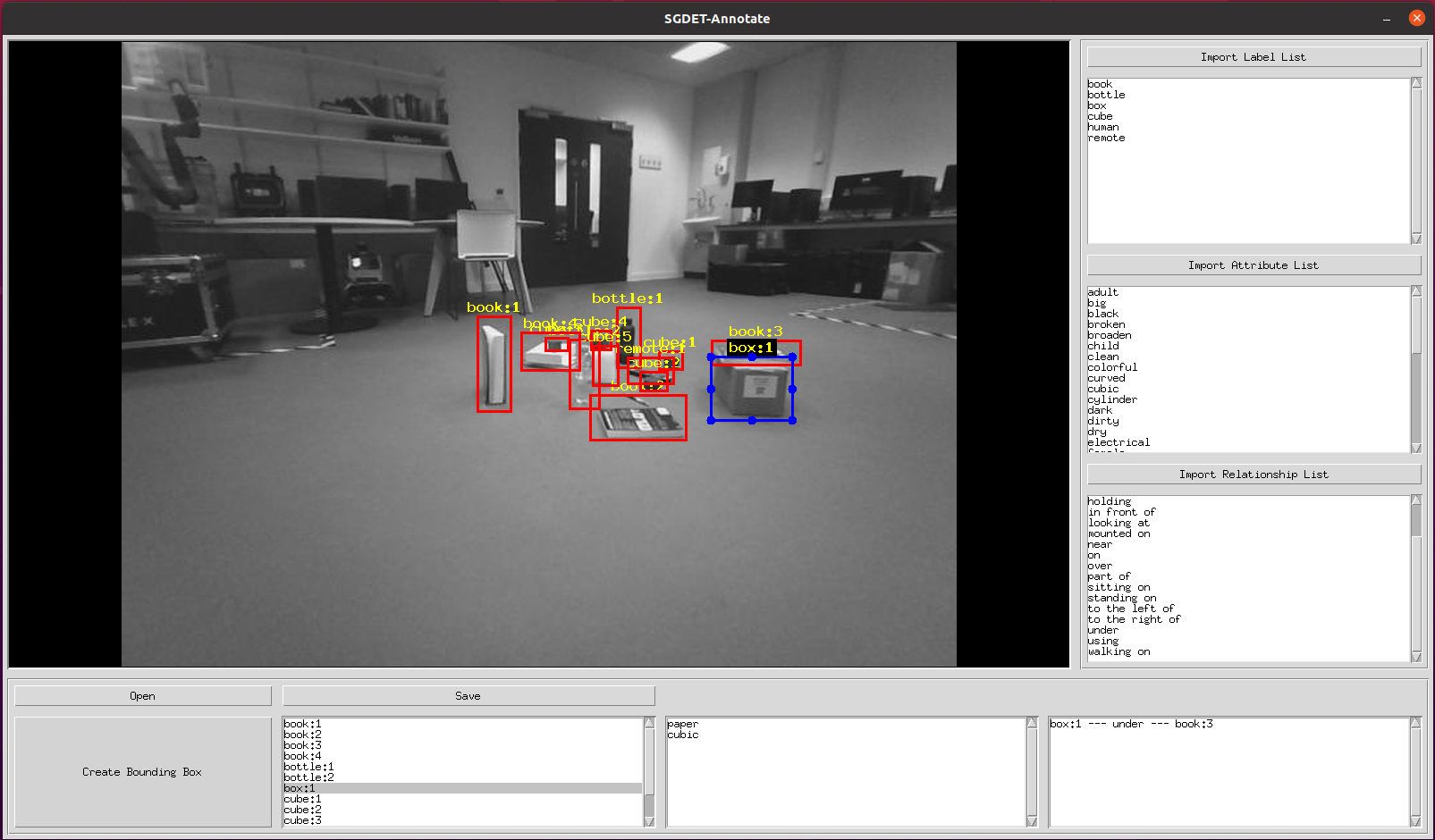}
    \caption{SGDET-Annotate (Please zoom in for the annotations): Our custom annotation tool for spatial relationship labelling. The left panel displays the robot-acquired image with editable bounding boxes; the right panel lists detected objects and available predicates. Annotators can select objects and predicates to efficiently create subject-predicate-object triplets. The tool exports annotations in both Visual Genome and YOLO formats for seamless model training and evaluation. SGDET-Annotate is open-sourced at \url{https://github.com/harvey-ph/SGDET-Annotate}.}   \label{fig:annotation_tool}
\end{figure}

\section{Evaluation of SGDet Predictors}

In this section, we evaluate the performance of several SOTA SGDet predictor models on our custom dataset of robot-acquired laboratory images. Our analysis focuses on inference speed, relational recall (both overall and per-predicate), and training convergence. By systematically comparing six leading architectures, we highlight their respective strengths, limitations, and the impact of dataset characteristics on their performance.

\subsection{Object Detection Backbone}

\begin{figure*}[ht]
  \centering
  \subfloat[Training losses over epochs\label{fig:yolo_train_loss}]{
    \includegraphics[width=0.3\linewidth]{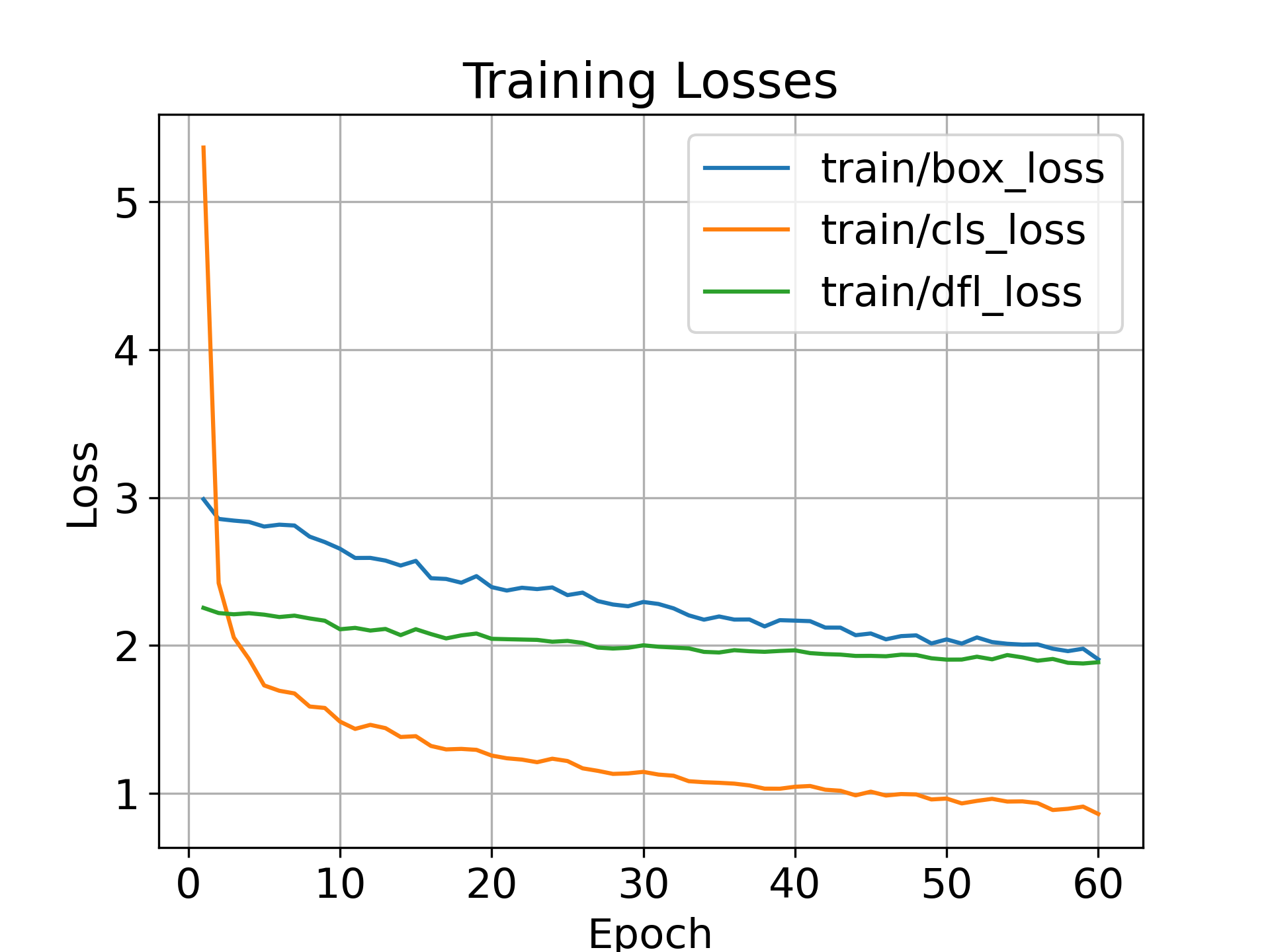}
  }
  \hfill
  \subfloat[Validation losses over epochs\label{fig:yolo_val_loss}]{
    \includegraphics[width=0.3\linewidth]{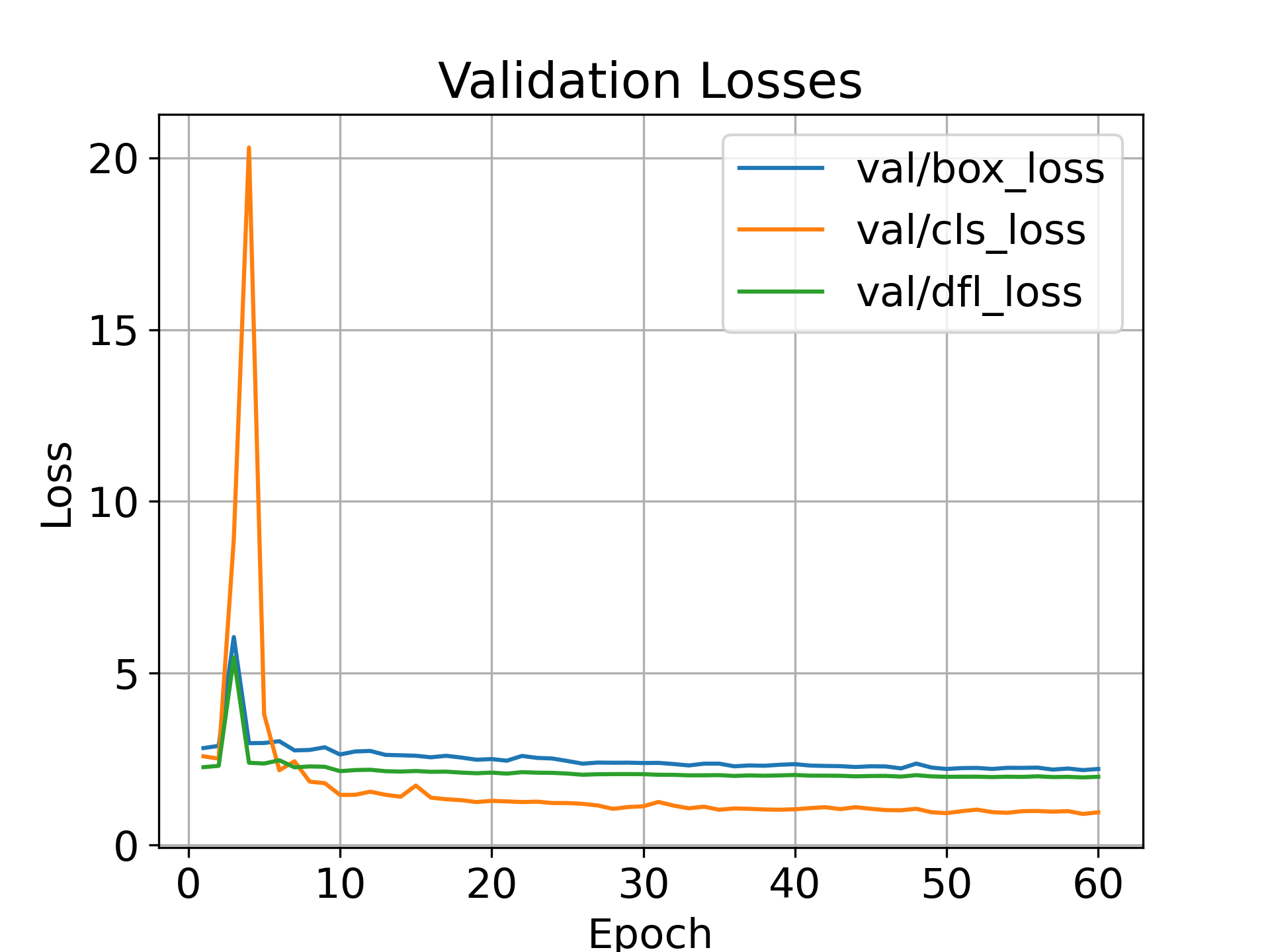}
  }
  \hfill
  \subfloat[Detection metrics over epochs\label{fig:yolo_metrics}]{
    \includegraphics[width=0.3\linewidth]{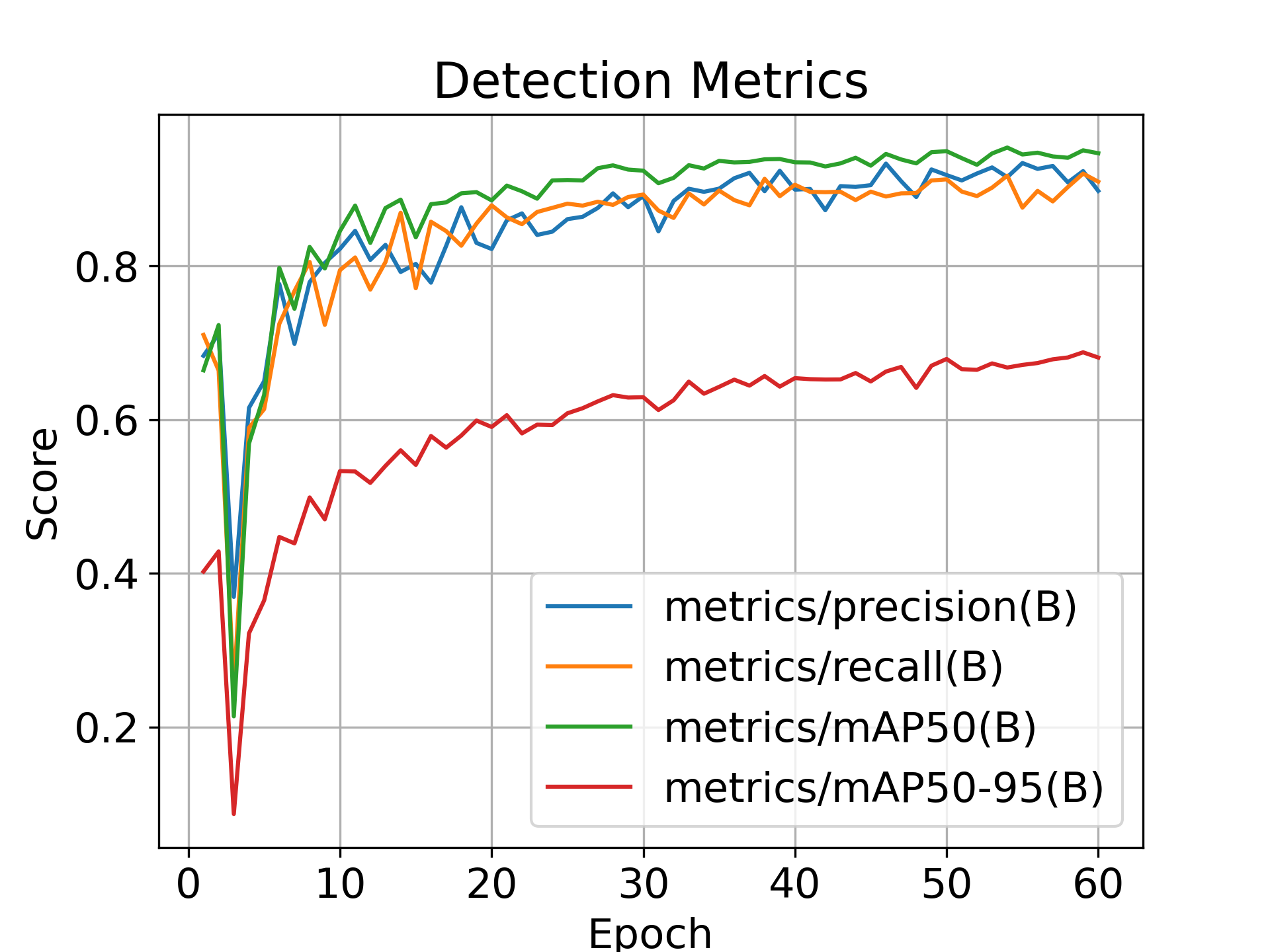}
  }
  \caption{Training results of YOLOv10m backbone over 60 epochs: (a) training losses, (b) validation losses, and (c) detection metrics. The box loss, classification loss, and distribution-focal-loss decrease steadily, indicating stable learning. The precision@50, recall@50, and mAP@50 metrics exceed 0.80 by epoch 10 and improve to approximately 0.92, 0.90, and 0.93, respectively, by epoch 60. The mAP@50-95 peaks around 0.68, reflecting the challenge of precise localisation at higher IoU thresholds. These results confirm that our YOLOv10m backbone converges quickly and provides a robust foundation for downstream relation-prediction evaluation.}
  \label{fig:yolo_results}
\end{figure*}

There are mainly two types of backbones to choose from, i.e., FasterRCNN and YOLO series. Following work from \cite{neau2024react}, we've chosen YOLOv10m due to its balance between accuracy and efficiency. Figures~\ref{fig:yolo_train_loss} and Figures~\ref{fig:yolo_val_loss} show the evolution of the three main loss components: box loss, classification loss, and distribution-focal-loss (dfl), on both the training and validation sets over 60 epochs. All losses decrease steadily, with the classification loss dropping sharply in the first 10 epochs and then tapering, while the box and dfl losses decline more gradually. The close tracking between training and validation losses indicates stable learning without overfitting.

Figure~\ref{fig:yolo_metrics} presents key detection metrics. By epoch 10, precision@50, recall@50, and mAP@50 all exceed 0.80, and continue to improve, reaching approximately 0.92 (precision), 0.90 (recall), and 0.93 (mAP@50) by epoch 60. The mAP@50-95 metric, which averages over all IoU thresholds, peaks around 0.68, reflecting the challenge of precise localisation at higher IoU. These results confirm that our YOLOv10m backbone converges quickly and provides a robust foundation for downstream relation-prediction evaluation.

\subsection{Inference Performance}

\subsubsection{Overall Latency and Aggregate Recall}

Table~\ref{tab:inference_performance} summarises the average and standard deviation of per-image inference latency for each model, along with retrieval metrics at K = 20, 50, 100 (R@K) and their macro-averaged counterparts (mR@K).

\begin{figure*}[!hbtp]
  \centering
  \includegraphics[width=1.0\linewidth]{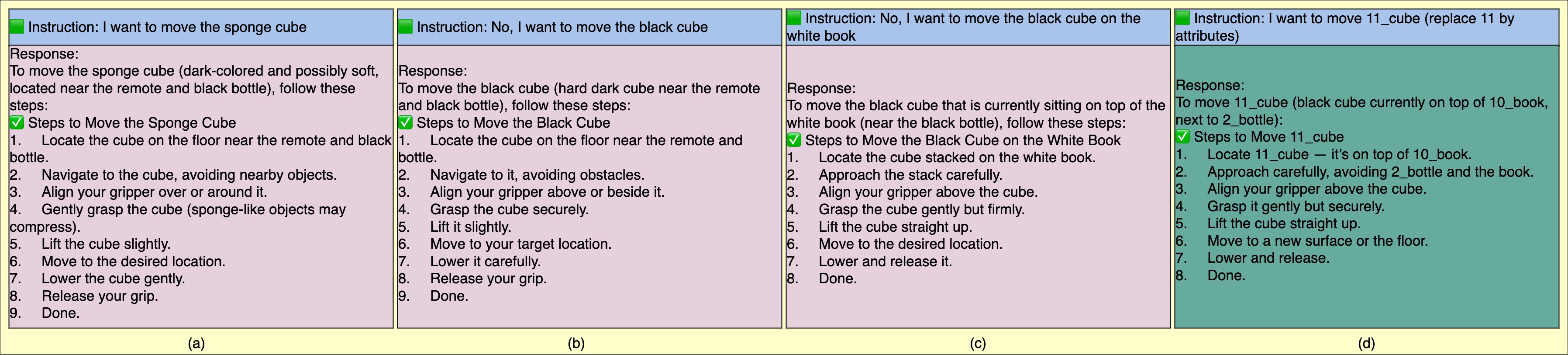}
  \caption{More examples of integrating SGDet results into ChatGPT 4o.}
  \label{fig:ChatGPT4o}
\end{figure*}

\begin{table*}
  \caption{Inference Latency and Retrieval Performance (R@K, mR@K) of Predictors}
  \label{tab:inference_performance}
  \begin{tabular}{lrrrrrrrrr}
    \toprule
    Predictor &Avg Latency (ms) &Std Latency (ms) &R@20 &R@50 &R@100 &mR@20 &mR@50 &mR@100\\
    \midrule
    Transformer Predictor & 27.5 & 18.6 & 0.3854 & 0.4415 & 0.4543 & 0.3956 & 0.4432 & 0.4556 \\
    VCTree Predictor & 92.5 & 21.4 & 0.4494 & 0.4792 & 0.4840 & 0.4576 & 0.4869 & 0.4909 \\
    Motif Predictor & 24.9 & 16.0 & 0.4277 & 0.4757 & 0.4856 & 0.4325 & 0.4736 & 0.4826 \\
    PE-NET Predictor & 36.5 & 21.9 & 0.3100 & 0.3294 & 0.3299 & 0.3065 & 0.3322 & 0.3327 \\
    Causal Analysis Predictor & 32.8 & 16.9 & 0.1414 & 0.1575 & 0.1575 & 0.1154 & 0.1289 & 0.1289 \\
    REACT Predictor & 45.4 & 18.1 & 0.1423 & 0.1561 & 0.1575 & 0.1165 & 0.128 & 0.1289 \\
    \bottomrule
  \end{tabular}
\end{table*}

Transformer Predictor and Motif Predictor achieve real-time performance (less than 30 ms per image) while maintaining competitive aggregate recall (R@100: 0.4543–0.4856). VCTree Predictor achieves the highest recall (R@100 = 0.4840; mR@100 = 0.4909) but at the cost of significantly higher latency. PE-NET offers moderate recall with higher latency, while Causal Analysis Predictor and REACT Predictor underperform in both speed and accuracy.

\subsubsection{Predicate-Wise Recall}

\begin{figure}
    \centering
    \includegraphics[width=1.0\linewidth]{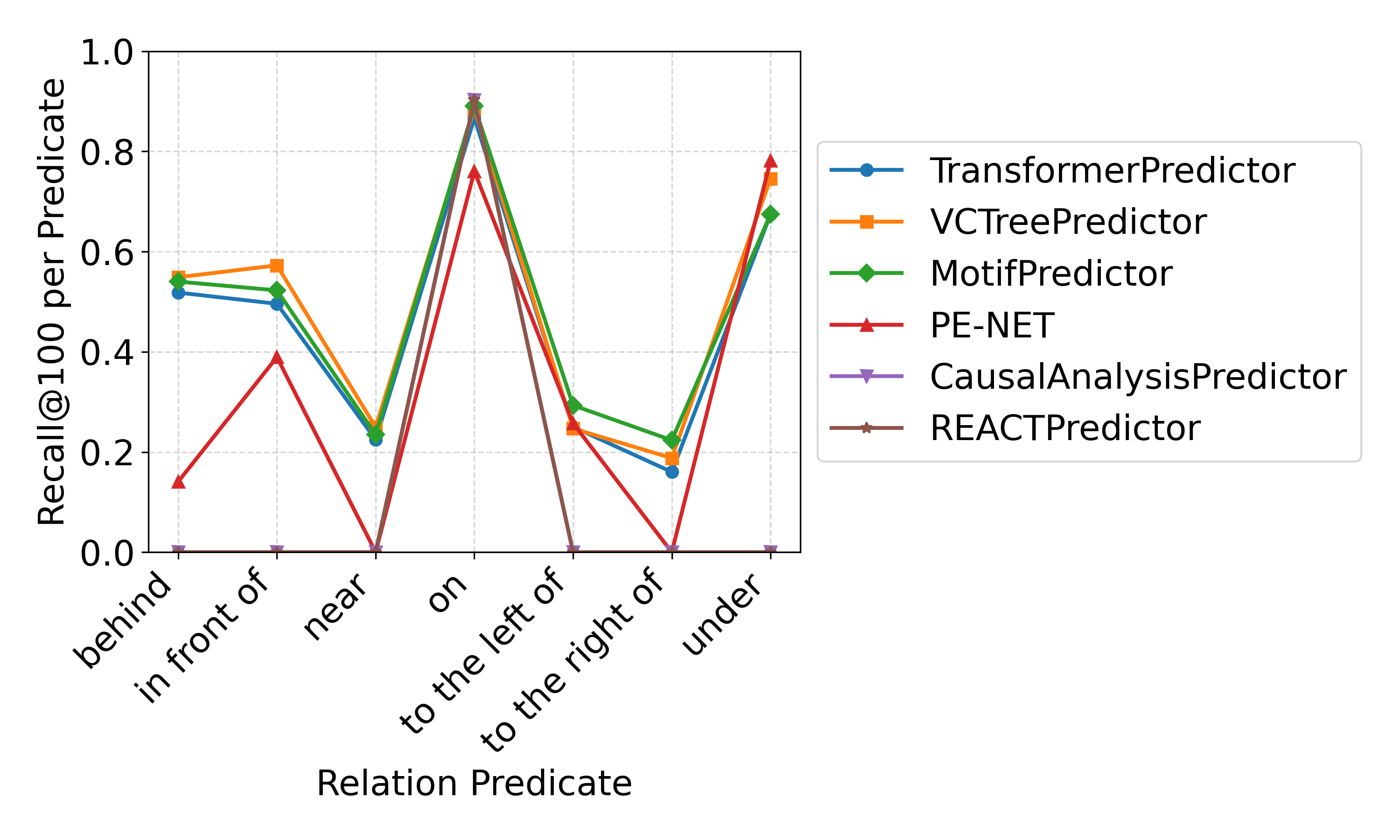}
    \caption{Per-predicate Recall@100 across predictors.}
    \label{fig:per-predicate}
\end{figure}

Figure~\ref{fig:per-predicate} shows Recall@100 per predicate for each predictor across the seven spatial relations in our dataset. All models achieve high recall on frequent relations such as \texttt{"on"} (0.7604–0.9025) and \texttt{"in front of"} (0.4957–0.5725), reflecting abundant training examples and strong visual cues. PE-NET performs best on \texttt{"under"} (0.7812). Lateral relations (\texttt{"to the left of"} and \texttt{"to the right of"}) show lower performance (0.1601–0.2931), and the \texttt{"near"} predicate remains challenging for all models (0.2247–0.2494). Causal Analysis Predictor and REACT Predictor fail to retrieve ground-truth instances for most relations. These results highlight the impact of label scarcity and annotation ambiguity on per-predicate performance.

\subsubsection{Training Convergence}

Despite a 50-epoch training schedule, all predictors reached their peak mR@100 well before the final epoch. Causal Analysis Predictor plateaued by epoch 5, REACT Predictor by epoch 2, and PE-NET by epoch 6. In contrast, Motif Predictor, VCTree Predictor, and Transformer Predictor required approximately 23–24 epochs to converge. This rapid saturation suggests that the dataset's limited diversity exhausts relational learning capacity early, making extended training less beneficial without additional data or curriculum strategies.

\subsection{Implications}

Our analysis demonstrates that dataset imbalance and annotation noise significantly affect per-predicate performance. Models with sophisticated context modelling (VCTree, Motif) achieve balanced recall across frequent predicates but still struggle with rare classes. PE-NET excels in vertical relations but underperforms on lateral ones, while causal modules are less effective in visually-defined tasks. To achieve robust scene-graph generation, future work should combine global attention and positional cues, employ early stopping based on predicate-wise mR plateaus, and augment under-represented relations while enforcing clear annotation guidelines (e.g., spatial thresholds for \texttt{"near"}).

Additionally, across all experiments, the YOLOv10m backbone maintained a consistent object-detection performance (mAP = 0.9086) on the validation set. This high and stable detection accuracy confirms that variations in relational recall and latency are primarily due to the design of the predicate-prediction heads, ensuring a fair comparison of downstream SGDet architectures.

\section{Integration to Foundation Models}
To further enhance the effectiveness of foundation models in robotic task planning, we integrated the spatial relationship information from our SGG models into ChatGPT 4o. This integration allows the model to leverage explicit spatial reasoning, improving its ability to generate executable plans for robotic tasks.
Figure~\ref{fig:teaser}f illustrates the process: the spatial relationships predicted by the SGG models are provided to ChatGPT 4o, which then generates a plan that incorporates this spatial reasoning. More results are given in Figure~\ref{fig:ChatGPT4o}, where we can see SGG models plus attributes will enhance VLM's performance in spatial awareness for robotic task planning.  Further demonstrates the potential of combining SGG models with foundation models to improve task execution in robotics.

\section{Conclusion}
In this work, we introduced a custom dataset of robot-acquired indoor images annotated with spatial relationships and object attributes, designed to benchmark the performance of SGG models in robotics contexts. Our dataset captures the complexities of real-world scenarios, including similar or identical objects and intricate spatial arrangements, reflecting the demands of robotic task planning.

We evaluated six SOTA SGG models, revealing significant differences in inference speed and relational accuracy. Our findings indicate that while some models excel in recall for frequent predicates, they struggle with rare classes, highlighting the impact of dataset characteristics on performance. Additionally, we demonstrated that integrating explicit spatial relationships into foundation models, such as ChatGPT 4o, significantly enhances their effectiveness in robotic task planning.

\begin{acks}
We thank the following MSc students for their efforts in annotating the dataset: Imad Ali, Nicholas Frederick, Pawel Rentz, Samaila Philemon Bala, and Thankgod Itua Egbe.
\end{acks}

\bibliographystyle{ACM-Reference-Format}
\bibliography{ACMSPATIAL}

\end{document}